\documentclass[conference]{IEEEtran}
\IEEEoverridecommandlockouts
\usepackage{cite}
\usepackage{amsmath,amssymb,amsfonts}
\usepackage{algorithmic}
\usepackage{graphicx}
\usepackage{textcomp}
\usepackage{xcolor}

\def\BibTeX{{\rm B\kern-.05em{\sc i\kern-.025em b}\kern-.08em
    T\kern-.1667em\lower.7ex\hbox{E}\kern-.125emX}}
\begin{document}

\title{Variations of Genetic Algorithms}

\author{Alison Jenkins, Vinika Gupta, Alexis Myrick, and Mary Lenoir}


\maketitle

\begin{abstract}

The goal of this project is to develop the Genetic Algorithms (GA) for solving the Schaffer F6 function in fewer than \begin{math}4000\end{math} function evaluations on a total of \begin{math}30\end{math} runs. Four types of Genetic Algorithms (GA) are presented - Generational GA (GGA), Steady-State \begin{math}(\mu+1)\end{math}-GA (SSGA), Steady-Generational \begin{math}(\mu,\mu)\end{math}-GA (SGGA), and \begin{math}(\mu+\mu)\end{math}-GA. 

\end{abstract}

\begin{IEEEkeywords}

genetic algorithm, elitism, generational, steady-state

\end{IEEEkeywords}

\section{Introduction}

Based on \begin{math}30\end{math} runs of the best performing EC variants (a total of \begin{math}12\end{math}), each crossover method for each type of GA is divided into its equivalent classes. One crossover method from the equivalence class (EC) that contains the smallest number of function evaluations is selected to represent the GA in a comparison of the four types of GAs. The three types of crossover methods studied are Single Point Crossover (SPX), Mid-Point Crossover (MPX), and Blend Crossover (BLX). SPX uses two parents to create two children using one cut point. MPX uses two parents to create one child. BLX uses two parents to create two children using one cut point. 

SPX uses two parents to create two children. SPX can be used for both binary coded and real coded representations. Each parent's chromosomes are divided into two parts by one cut point. The first part of the first parent is combined with the second part of the second parent to create the chromsome for the first child. The second child's chromosome is then created by the combination of the second part of the first parent and the first part of the second parent.

MPX uses two parents to create one child. MPX is used only for real-coded representations. Given the two floating point numbers representing the genes of each parent, the midpoint between these two parent genes is calculated simply by adding them and dividing by two. This calculated midpoint is then assigned as the gene for the child. This process repeats for each gene in the length of the chromosome.

BLX uses two parents to create one offspring. BLX is also used for real-coded representations. Given two floating point numbers for each of the parent genes, the child will be assigned a random value within the range between these numbers. This process repeats for each gene in the length of the chromosome.

SPX and MPX do not select very well, because they do not introduce much variation. 

This paper highlights the comparison between the generational, steady state, steady generational and \begin{math}(\mu + \mu)\end{math} genetic algorithms. It gives a detailed comparison by depicting the performance of each algorithm with all 3 above mentioned crossovers, i.e. SPX, MPX and BLX.

\section{Methodology}

Binary Tournament Selection, with recombination (crossover) operators: SPX, MPX \begin{math}(p_{mp}{=}1.0)\end{math}, and BLX-\begin{math}0.0\end{math}, is studied in this project. An analysis of each crossover method for each type of GA (GGA, SSGA, SGGA, and \begin{math}(\mu+\mu)\end{math}-GA) evaluates the number of function evaluations of each method. Equivalence classes are determined based on the ANOVA and Student T-test results.

\subsection{Generational GA}

In the GGA, if the population size is P, there are P offspring that are created and mutated. Following this, the replacement strategy replaces all the parents with their offspring. This results in no overlap between the current and new population. In this case, elitism is \begin{math}0\end{math}. Here, the generation gap, which is the measure of degree of overlap between the current and new generation, is zero. 

The GGA uses the tournament selection method to select the two parents from the population to create one offspring. For each offspring, two parents are randomly selected from the existing population, and the process repeats until the number of offspring reaches the current population size. Every member of the current population is eliminated, and zero survivors remain.  

\subsection{Steady-State GA}

The SSGA works by randomly selecting two parents, creating one offspring, and replacing the worst fit individual in the population with the offspring. 

The benefit of using a SSGA, rather than a GGA, is that the SSGA makes only one function evaluation per child on each cycle. A GGA must make P (where P is the population size) function evaluations on each cycle. Therefore, function evaluations are only counted when comparing GGAs with SSGAs.

\subsection{Steady-Generational GA}

The SGGA works by first selecting the two parents and then generates the offspring. Instead of the offspring replacing the parents or the worst-fit individual, the offspring replaces a random individual from the population that is not the best-fit.

Similar to the SSGA, the SGGA has the benefit of only making two function evaluations on each cycle. The GGA requires P function evaluations on each cycle.

\subsection{$(\mu+\mu)$-GA}

The {\begin{math}(\mu+\mu)\end{math}-GA works by randomly selecting two parents with binary tournament selection, creating an offspring, and adding the offspring to a child population until the child population size is equal to the original population size. The algorithm then creates a new population containing the original population and the child population, and chooses the top individuals from this new population until the population size is the same size as the original population. In this way, the algorithm composes a population of the most fit individuals out of two generations of individuals. 

The benefit of using the \begin{math}(\mu+\mu)\end{math}-GGA is that although there are more function evaluations each cycle, the best fit individuals are guaranteed in the new population as opposed to randomly replacing individuals and potentially ending up with a a population with lower fitnesses. 

\section{Experiment}

An upper bound of \begin{math}100\end{math} and a lower bound of -\begin{math}100\end{math} is used in both x and y dimensions, and a \begin{math}F1\end{math} score is used to determine the fit of each member of the population. An optimal genetic algorithm with a population size of \begin{math}16\end{math} and a mutation rate of \begin{math}0.012\end{math} is used. 

The average number of function evaluations is found for \begin{math}30\end{math} runs of each GA combined with each crossover method. In statistics, \begin{math}30\end{math} runs is typically considered acceptable for determining the average of algorithms and performing the ANOVA and Student T-tests to determine equivalence classes.

\subsection{ANOVA Test}

The ANOVA test is an analysis of variance that is used to determine if a statistically significant difference exists in the performance of the various GAs. 

If the p-value given for each combination of GA variations is smaller than \begin{math}p < 0.05\end{math}, then the variances differ such that there is a statistically significant difference between the two algorithms. 

Therefore, the number of algorithms in the analysis is reduced by one. Then, the ANOVA test is run again on the remaining algorithms. The previous two steps are repeated until the number of algorithms results in a value for p that is greater than 0.05. A Student T-test (1-tailed or 2-tailed) is run to determine the variance. 

Microsoft Excel provides both a T-test and ANOVA test through the Data Analysis Toolbox Add-In to determine equal or unequal variances. The \begin{math}F1\end{math} algorithm may be used to determine whether to use the ANOVA or Student T-test.

\subsection{Student T-test}

If \begin{math}abs(t) > 1.7\end{math}, then there is a statistically significant difference in value between the two algorithms. (Even though their average values are different.) 

If \begin{math}abs(t) = 1.5\end{math}, then the two algorithms are in the same equivalence class. 

If \begin{math}abs(t) = 1.9\end{math}, then the two algorithms are in different equivalence classes.

\subsection{Generational GA}

With the population size, P of 16 and a mutation amount of 0.012, 2p parents are selected at random to create p offspring with a selection pressure of 0. Then, SPX is used to obtain the next generation of p offsprings. The results are noted for 30 consecutive runs of maximum evaluations. This process is repeated for MPX and BLX. Then, these results are compared using ANOVA and Student T-tests.

\subsection{Steady-State {$(\mu+1)$}-GA}

The Student T-test results in the SPX, MPX, and BLX variations on the SSGA algorithm being in the same equivalence class. The average number of function evaluations is 1497 for the MPX SSGA, which is the best algorithm. The ANOVA and Student T-tests place all of the SSGA crossover methods in the same equivalence class.

\subsection{Steady-Generational {$(\mu,\mu)$}-GA}

The SGGA results in the best algorithm, the MPX SGGA, having an average value of function evaluations of 2636.

\subsection{{$(\mu+\mu)$}-GA}

The \begin{math}(\mu+\mu)\end{math}-GA's MPX recombination operator has the best average for function evaluations out of the recombination operators for this GA, and therefore the ANOVA test and Student T-test includes the MPX recombination operator for the best performing GAs. The ANOVA test is performed for all four of the best GAs and results in a very small p-value, indicating that the variances differs. When the ANOVA test is performed for the \begin{math}(\mu+\mu)\end{math}-GA and the GGA, a p-value greater than 0.05 is found, meaning the variances of these two algorithms are similar. To determine what Student T-test to use, an F-Test is performed in order to compare the mean value of the two algorithms, and to compare the F and F Critical values. The T-Test Assuming Equal Variances is used since the mean value of GGA is less than the mean value of \begin{math}(\mu+\mu)\end{math}-GA, and the F value is greater than the F Critical value. 

Performing the Student T-test assuming equal variances results in a t-Stat value of less than 1.7. This confirms that the mean difference is zero for the two GAs. Therefore, the GGA and the \begin{math}(\mu+\mu)\end{math}-GA are in the same equivalence class. 
\begin{figure}
\begin{center}
\setlength{\unitlength}{0.012500in}
\includegraphics[width=75mm, scale=0.8]{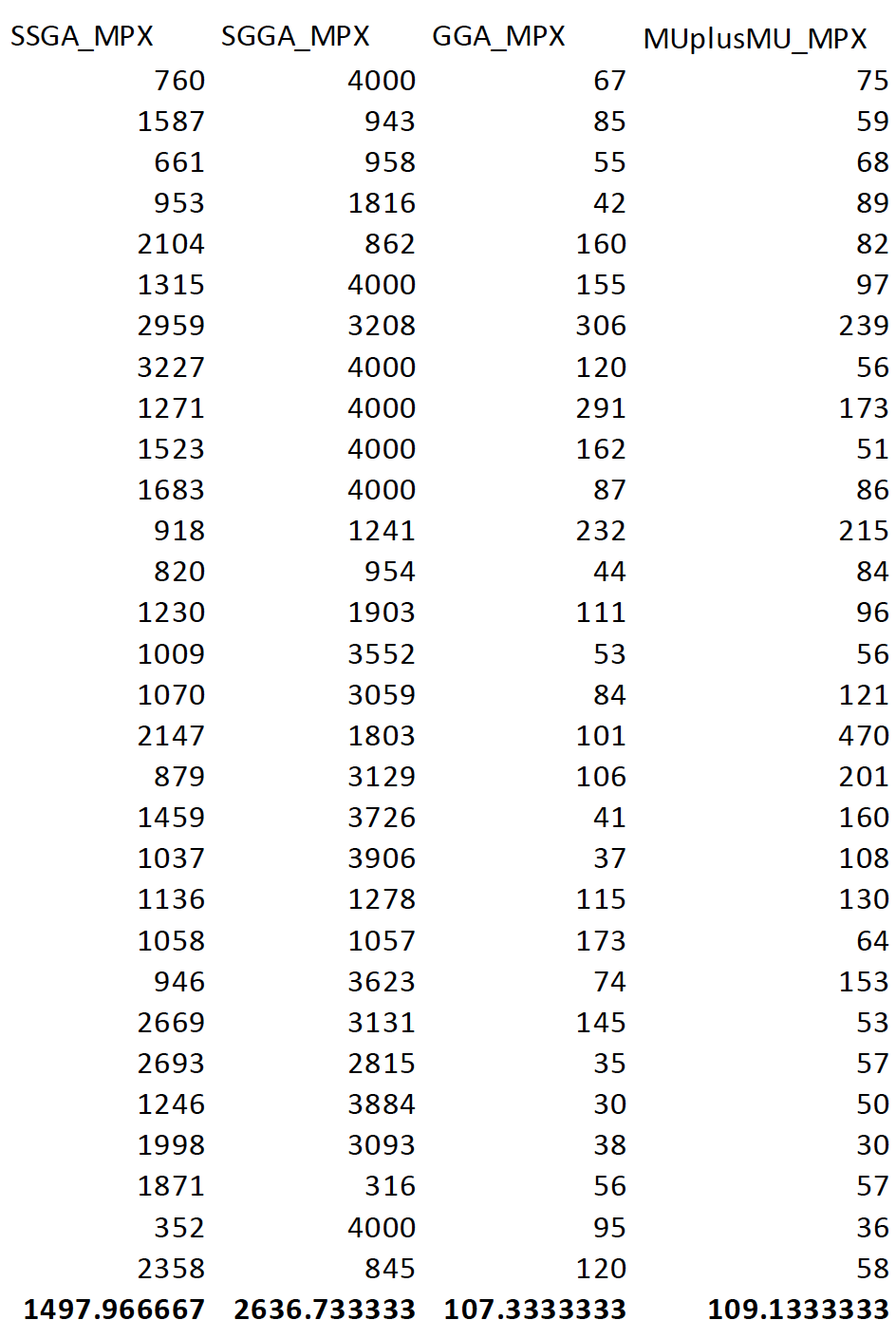}
\end{center}
\caption{GA Data Set}
\label{figureGA_DATA} 
\end{figure}

\begin{figure}
\begin{center}
\setlength{\unitlength}{0.012500in}
\includegraphics[width=75mm, scale=0.8]{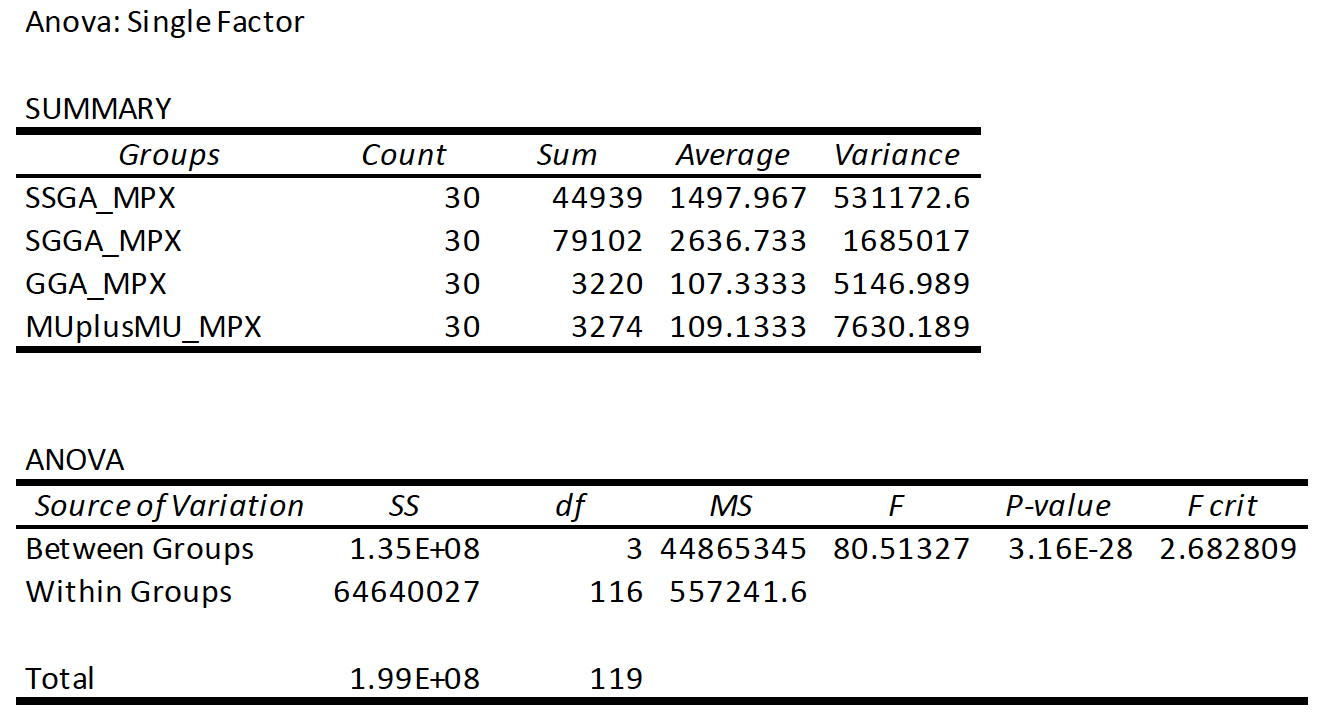}
\end{center}
\caption{ANOVA Results}
\label{figureGA_ANOVA} 
\end{figure}

\begin{figure}
\begin{center}
\setlength{\unitlength}{0.012500in}
\includegraphics[width=75mm, scale=0.8]{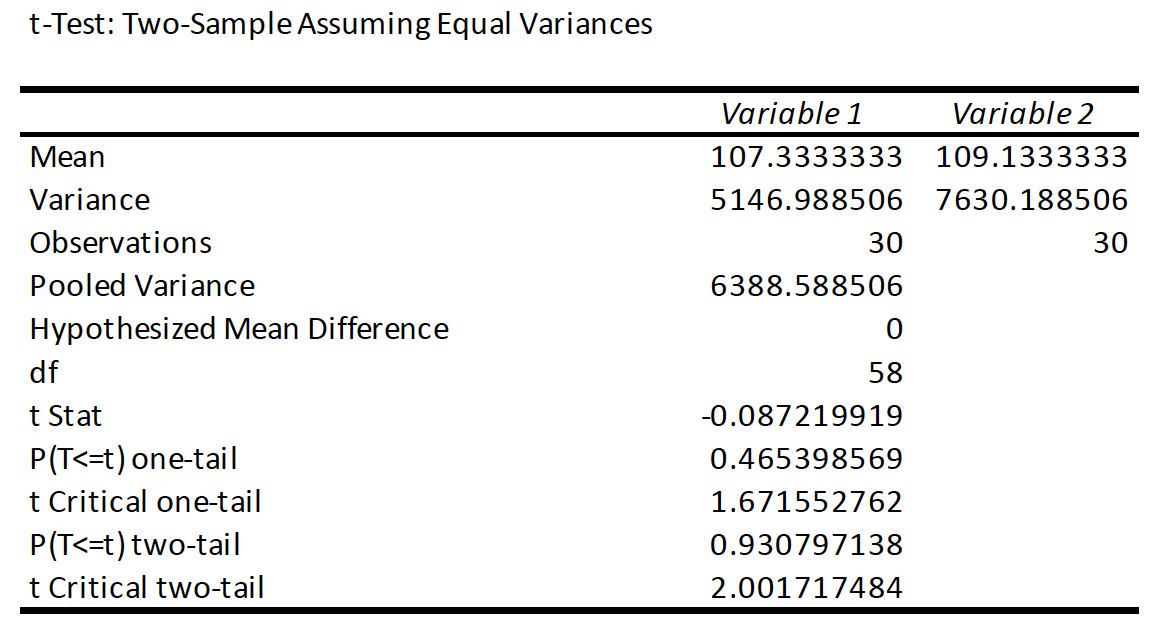}
\end{center}
\caption{Student T-test Results}
\label{figureGA_Ttest} 
\end{figure}

\section{Results}




The results from each crossover method for the SSGA places all of the methods in the same equivalence class using both the ANOVA and Student T-tests. The data set used is shown in Figure~\ref{figureGA_DATA}, while the ANOVA test results are shown in Figure~\ref{figureGA_ANOVA}. The results of a Student T-test are shown in Figure~\ref{figureGA_Ttest}. 


The results from the crossover operators for the \begin{math}(\mu+\mu)\end{math}-GA assigned a P-value of 0.78 when the ANOVA test is performed, which places all of the crossover operators in the same equivalence class. The Student T-Tests for each pairing of the crossover operators also all had t-Stat values with an absolute value less than 1.7, indicating that there is no statistically significant difference between each of the pairs of crossover methods.

For the ANOVA test with each of our crossover operators resulting in the least average number of function evaluations, a very small P-value was obtained which indicates no statistically significant difference in our four algorithms. However, the ANOVA test for the {\begin{math}(\mu+\mu)\end{math}-GA and GGA resulted in a P-value of 0.93, meaning that these two algorithms are in the same equivalence class. The Student T-Test resulted in a t Stat of -0.08722, so the null hypothesis is accepted. The null hypothesis is that the hypothesized mean difference is zero. The GGA MPX, which is the algorithm with the better mean, is a statistically better algorithm on this problem instance.

Each GA provides unique benefits and challenges to solving a problem. In comparing the various GAs, the number of function evaluations differs based on the number of evaluations needed for each cycle. 

The average number of function evaluations for \begin{math}30\end{math} runs of each GA combined with each crossover method for an optimal genetic algorithm with a population size of \begin{math}16\end{math} and a mutation rate of \begin{math}0.012\end{math} is found. The GA algorithms are compared using the ANOVA and Student T-tests to find the combination with the smallest number of function evaluations and the equivalence classes are determined. The SGGA MPX is the algorithm with the worst average out of the other GAs, with an average value of 3636.73 function evaluations. The GGA MPX is found to have the lowest function evaluations and is in the same equivalence class as  \begin{math}(\mu+\mu)\end{math}-GA MPX. These perform considerably better than the SSGA MPX and SGGA MPX.

\section{Breakdown of the Work}

Vinika Gupta - Generational GA, Analysis. 

Alison Jenkins - Steady-State GA, Analysis, and LaTeX Report. 

Alexis Myrick - Steady-Generational GA, Analysis. 

Mary Lenoir - \begin{math}(\mu+\mu)\end{math}-GA, Analysis.

\vspace{12pt}

\begin{thebibliography}{00}
\bibitem{b1}{Engelbrecht, Andries P. \textit{Computational Intelligence: An Introduction}. John Wiley \& Sons, 2007.}
\bibitem{b2}{Joseph, Anthony D., et al. \textit{Adversarial Machine Learning}. Cambridge University Press, 2018.}
\bibitem{b3}{Sarkar, Dipanjan. \textit{Text Analytics with Python}. Apress, 2016.}
\bibitem{b4}{Dozier, J. \textit{Genetic Algorithms}. Powerpoint Presentation, COMP6970 - Computational Intelligence and Adversarial Machine Learning class, Auburn University, 2019.}
\end{thebibliography}
\end{document}